\documentclass{article}

\usepackage[preprint]{neurips_2021}

\usepackage[utf8]{inputenc} 
\usepackage[T1]{fontenc}    
\usepackage{hyperref}       
\usepackage{url}            
\usepackage{booktabs}       
\usepackage{amsfonts}       
\usepackage{nicefrac}       
\usepackage{microtype}      
\usepackage{xcolor}         
\usepackage{natbib}
\usepackage{siunitx}
\usepackage{graphicx}
\graphicspath{ {./images/} }

\title{Generating coherent comic with rich story using ChatGPT and Stable Diffusion}

\author{%
  Ze Jin\\
  University of Toronto\\
  \texttt{ze.jin@mail.utoronto.ca} \\
  \And
  Zorina Song\\
  University of Toronto\\
  \texttt{zo.song@mail.utoronto.ca} \\
}

\begin{document}

\maketitle

\begin{abstract}
Past work demonstrated that using neural networks, we can extend unfinished music pieces while maintaining the music style of the musician. With recent advancements in large language models and diffusion models, we are now capable of generating comics with an interesting storyline while maintaining the art style of the artist. In this paper, we used ChatGPT to generate storylines and dialogue and then generated the comic using stable diffusion. We introduced a novel way to evaluate AI-generated stories, and we achieved SOTA performance on character fidelity and art style by fine-tuning stable diffusion using LoRA, ControlNet, etc.
\end{abstract}

\section{Introduction}
Previously, Huawei created a model that extended Franz Schubert’s Symphony No. 8, a piece famously left unfinished.\citep{huawei}  This is a good idea that can be applied to not just music, but also comics. There are lots of incentive to be able to extend a comic, it can be done when the author left the comic unfinished, it can be done in case people doesn't like the ending of a particular comic, etc. In this paper, we picked the popular Japanese Manga One Piece as the target to extend and evaluate performance on, but the method we used can be applied to any comic.

\section{Related Work}

\subsection{GPT}
OpenAI recently released its latest GPT-4, it outperformed the previous benchmark GPT-3.5 in a variety of exams according to OpenAI.\citep{openai2023gpt4} But there is no existing work testing GPT-4 ability to generate stories, so we will be comparing GPT-3.5 against GPT-4.

\subsection{Diffusion Models}
With story generation handled by GPT, the next step is to pick a model that draws the comic. Midjourney\citep{midjourney} is the current SOTA choice for generating comics, the results are very high fidelity\citep{Katz2022}, however, Midjourney is closed-sourced and accessible only via cloud services. Therefore, we will be working on Stable Diffusion\citep {compvis}, where we can fine-tune the model.

\subsection{Visual ChatGPT and HuggingGPT}
The plan of this paper is to combine ChatGPT and Stable Diffusion, where we used the output of ChatGPT and feed it into Stable Diffusion. Visual ChatGPT\citep{wu2023visual} and HuggingGPT\citep{shen2023hugginggpt} already did that, but because they are using stable diffusion base models, they can't generate Characters from our target Manga One Piece.

\subsection{Dreambooth, Textual Inversion, LoRA - Low-Rank Adaptation}
If stable diffusion base model is not capable of generating characters from One Piece, then we need fine-tuning. LoRA\citep{hu2021lora} is our choice for fine-tuning stable diffusion because LoRA offers a good trade-off between file size and training power. Dreambooth\citep{ruiz2023dreambooth} is powerful but results in large model files (2-7 GBs). Textual inversions\citep{gal2022image} are tiny (about 100 KBs), but you can’t do as much. The LoRA we will be using in this paper is Wano Saga made by \citep{lykon}.

\section{Methods}

\subsection{Generating comic story with ChatGPT}
We used ChatGPT to generate one page of the comic at a time, where each page of the comic contains 6 panels, and each panel contains a scene description and dialogue between one or two characters. The prompt we used is: "Write a comic book page with six comic book panels with descriptions and dialogs. Characters can only be Nami, Zoro, Monkey D. Luffy, and only one or two characters per panel." To reduce the amount of work, we told ChatGPT to limit the story to contain only three characters, Luffy, Zoro, and Nami. Then, the prompt "continue for another page" is used to generate another page.

\subsubsection{Generating comic story with ChatGPT with extra biography prompt}
Because we are not sure how much ChatGPT knows about One Piece, we wanted to provide knowledge of One Piece to ChatGPT. Therefore, we also tried entering the biography of Luffy, Zoro, and Nami that we got from One Piece Wiki\citep{fandom} as three different prompts to ChatGPT prior to the actual comic generation prompt.

\subsection{Calculate Story Score}
Since there is no existing metric that evaluates AI-generated stories, we will create one. Introducing story score, the story score of story $x$ against target story $t$ and popular Manga set M:

\begin{equation}
story(x ; t, M) = \gamma sim(x, t) + (1-\gamma) plot(x; M)
\end{equation}
where $sim(x, t)$ is the similarity score of story $x$ against the plot of the target Manga One Piece, and $plot(x)$ is the plot score of story $x$ against the story of a collection of popular Manga $M$,
\begin{equation}
plot(x; M) = \frac 1 {|M|} \sum_{m \in \mathcal{M}}sim(x, M)
\end{equation}
with originality adjusting factor $0 \le \gamma \le 1$.

The reason behind the story score is, firstly, we need the newly generated story to be a continuation of the target story, therefore we want it to have some similarity with the original story. Then, we also want the generated story to be good, therefore we compare it against a set of popular Manga $M$, we calculate the plot score as average similarly score on the set of popular Manga $M$.

In this paper, we calculated a simplified version of the story score to reduce the amount of computation, where we calculate the story score using the summary of ChatGPT generated story against the summary of the target story, and the summary of popular Manga set $M'$.
\begin{equation}
story(x; t, M') = \gamma sim(smry(x), smry(t)) + (1-\gamma) plot(smry(x); M')
\end{equation}
We summarized the ChatGPT generated story using ChatGPT, because according to this \citep{chen} and other articles, GPT is currently the best option for text summarization. Then we calculated the similarity score using pre-trained BERT model\citep{devlin2019bert}, with sentence-encoder\citep{pypi}.  We get the summary's embedding using BERT model, then calculate the cosine similarity of the embeddings, our code is on Github\citep{song_jin}. The popular Manga set we compared against is

$M' = \{Naruto, Bleach, Hunter X Hunter, Attack On Titan, Fullmetal Alchemist\}$

And we used $\gamma = 0.5$.

\subsection{Generating Character with Stable diffusion}
The scene description generated using ChatGPT contains the character, background, and action. Our hypothesis is that, before we generate the entire scene, we should fine-tune the model so that it can generate the characters first, once our model is capable of generating characters, then we will ask it to draw the entire scene which contains the characters along with the background and actions.

\subsubsection{Generating Character - Checkpoints/base models}
To use Stable Diffusion, we can either start with a base model like v2.1, or checkpoints like waifu-diffusion\citep{hakurei} and Midnight Mixer Melt\citep{drbob2142} that already had fine-tuning done on anime images.

\subsubsection{Generating Character - Fine-Tuning}
Then, we came up with a sequence of steps to fine-tune stable diffusion models. First, apply LoRA made by \citep{lykon} that was specifically fine-tuned on One Piece characters. Next, adjust the parameters, Sampling Steps from 20 to 30, CFG Scale from 7 to 9, and sampler from Eucla A to DPM++2M Karras. Next, change Clip-Skip\citep{elnouby2019skipclip} to 2 because our base model was trained to perform better on Clip-Skip. Next, use waifu-
diffusion-v1-4's VAE \citep{kingma2022autoencoding} made by \citep{hakurei} that was trained on anime images to produce better color. Then, add extra prompts with keywords "wanostyle", "solo", "stand straight", "long green clothes", "white clear background". Lastly, the character already looks good by this point, but the pose of characters still varies based on the random seed, to have the character in the exact pose we want, use ControlNet\citep{zhang2023adding} with Canny model\citep{lllyasviel}. Where Canny's pre-processor extracts the pose in any given image, and Canny's model makes stable diffusion generate character based on the given pose.

\subsection{Calculating SSIM and FID}
We used SSIM\citep{1284395} and FID\citep{heusel2018gans} to evaluate the result after fine-tuning stable diffusion against the target. To calculate SSIM, we feed the result and target batch into Pytorch's conv2d with a filter sampled from a Gaussian distribution, then used the output's mean and variance to compute SSIM. To calculate FID, we feed the result and target batch into Keras' InceptionV3, then we used the output's mean and variance to compute FID, our code is on Github\citep{song_jin}.

\section{Experiment}
\subsection{Generating comic story}
We generated comic stories using GPT-3.5 and GPT-4 through ChatGPT, with and without extra character biography prompts, then calculated story scores for each story using the method in 3.2.

\subsection{Generating Character}
We picked the character Roronoa Zoro from One Piece as the target, we used 5 original images of Rorona Zoro as the target batch. We posted the target batch we used in the appendix.

\subsubsection{Generating Character - Checkpoints/base models}
The first step is to evaluate checkpoints/base models. We evaluated Midjourney V4, Midjourney V5, stable diffusion v2.1, and Midnight Mixer Melt, on their ability to generate Roronoa Zoro from One Piece. We used the default parameters, with the bare minimum prompt "Roronoa Zoro, One Piece, Katana, Full body". For each model, we generated a batch of 5 images of Roronoa Zoro using 5 random seeds. And then we used the batch of 5 images generated by each model to calculate SSIM and FID scores against the batch of 5 target images.

\subsubsection{Generating Character - Fine-Tuning}
Then we picked Midnight Mixer Melt to fine-tune and evaluated the effect of each fine-tune step. In order to fairly evaluate the effect of each fine-tuning step, we fixed 5 random seeds to be reused in every step. We performed the sequence of fine-tuning steps in 3.3.2, and used the batch of 5 outputs to calculate the SSIM and FID against the target batch after each fine-tuning step.

\section{Results}
\subsection{Generating Story - Story Score}
\begin{table}[h]
\centering
\begin{tabular}{SSSSSSSS}
\toprule
    {Model} & {Similarity Score} & {Plot Score} & {Story Score} \\ \midrule
    {GPT-4 with biography prompt} & \textbf{ 0.57} & \textbf{ 0.67} & \textbf{ 0.62}  \\
    {GPT-4  without biography prompt}  & 0.49  & 0.66 & 0.57 \\
    {GPT-3.5 with biography prompt}  & 0.52  & 0.58 & 0.55 \\
    {GPT-3.5 without biography prompt}  & 0.54  & \textbf{ 0.67} & 0.6   \\ \bottomrule
\end{tabular}
\end{table}
Giving extra prompts and extra knowledge about One Piece does not necessarily generate a better story both by looking at the story score and reading the story by humans. Also, GPT-3.5 and GPT-4 performed similarly both by looking at the story score and reading the story by humans. But overall, GPT-4 with extra biography prompt performed slightly better. We posted sample stories that we generated in the appendix.

\subsection{Generating Character - SSIM and FID Score}
\includegraphics[width=\textwidth]{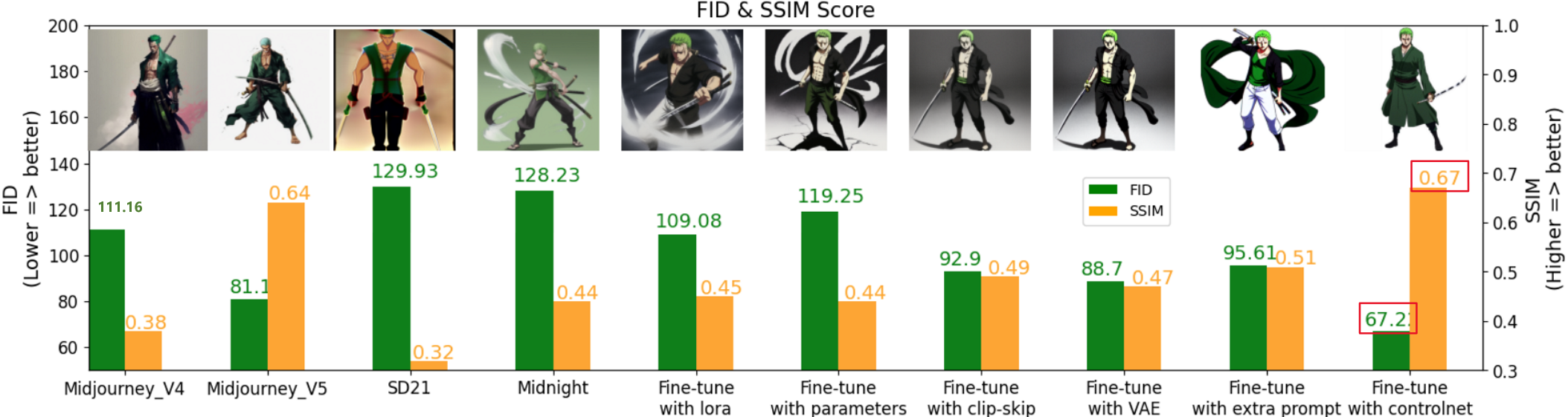}
Midjourney V5\citep{midjourney} initially performed the best out of all the checkpoints/base models. But in the fine-tuning steps, our result gradually improves for each step both by looking at the score, and in human eyes. Then, after all the fine-tuning steps, our model did outperform the benchmark Midjourney V5. We posted the images generated  before and after fine-tuning in the appendix.

\subsection{Comic Results}
We don't have space to evaluate the result of generating full comics for this paper, but it does look like our hypothesis from 3.3 holds true, we posted sample comics that we generated in the appendix.

\section{Conclusion}
In conclusion, we found a good way to generate comic scenes with ChatGPT and evaluated the story using the story score we introduced, then we found an effective sequence of steps to fine-tune stable diffusion to generate the character in the target Manga and outperformed Midjourney V5.
\subsection{Limitation and Future work}
For comic characters, our experiment could be better if we can test more characters and larger random batches. Also, SSIM and FID score is not the perfect way to evaluate the AI-generated comic, we don't have better metrics. For the story, our story score metric needs more testing and proof. Due to model and compute limitations, we calculated story scores using stories summarized by ChatGPT.

\section{Appendix}
\subsection{Sample Character - Roronoa Zoro from One Piece}
\subsubsection{target batch}
\includegraphics[width=\textwidth]{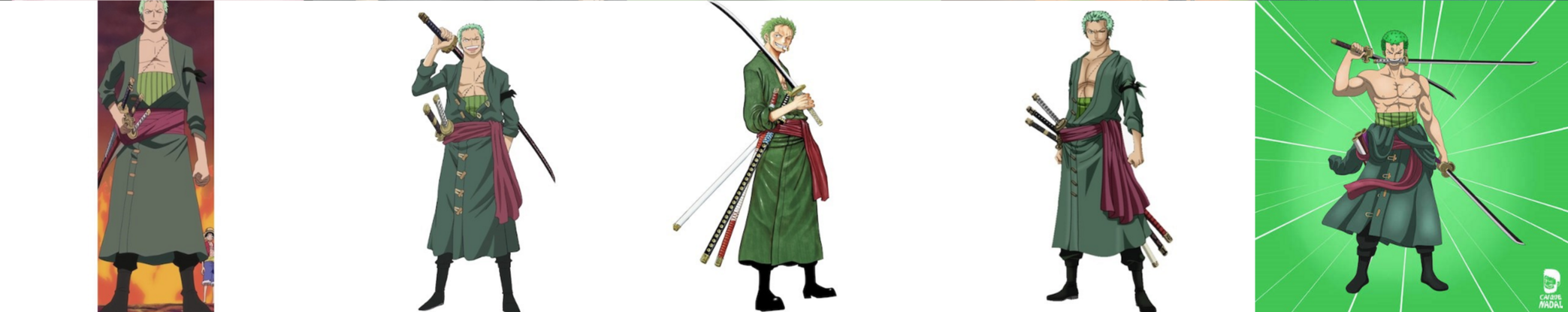}
\subsubsection{Midjourney V5 benchmark}
\includegraphics[width=\textwidth]{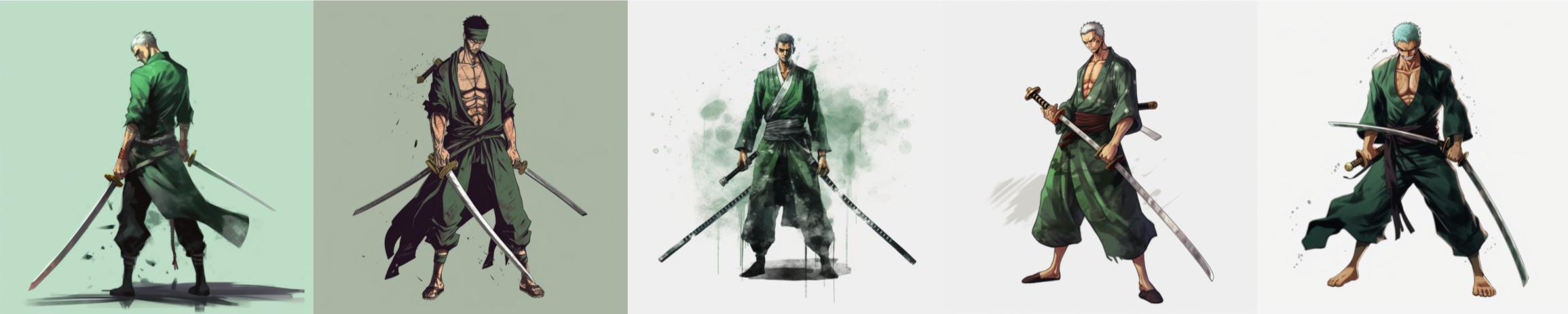}
\subsubsection{before fine-tuning}
\includegraphics[width=\textwidth]{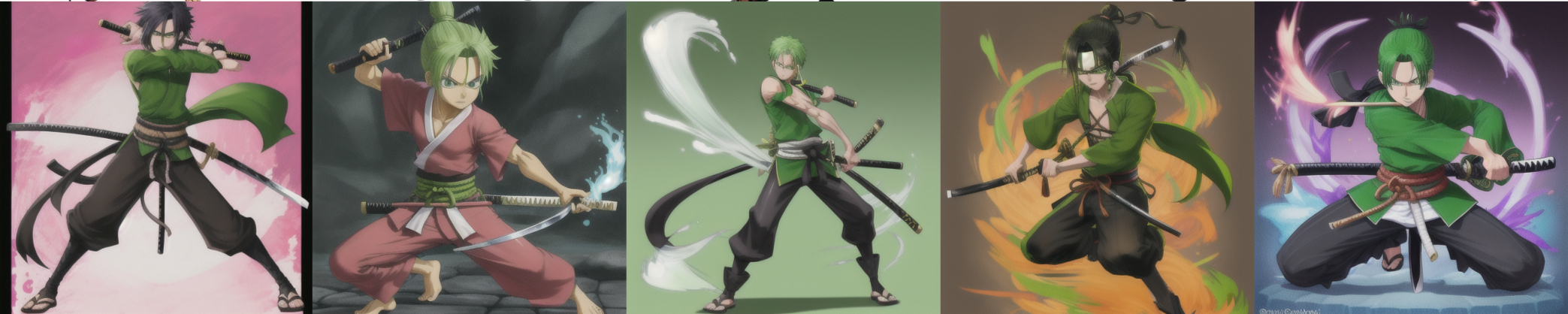}
\subsubsection{after fine-tuning}
\includegraphics[width=\textwidth]{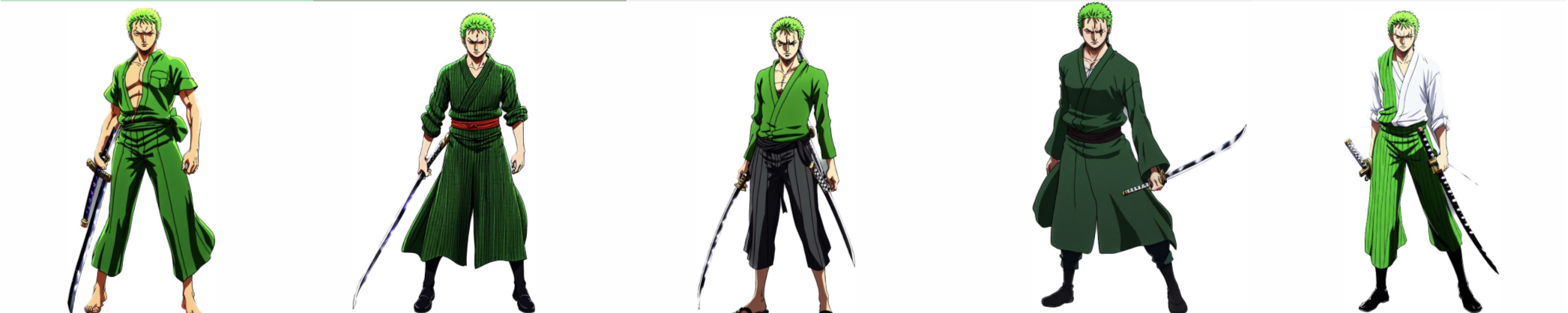}
\subsection{Sample Story Generated using ChatGPT}
\includegraphics[width=\textwidth]{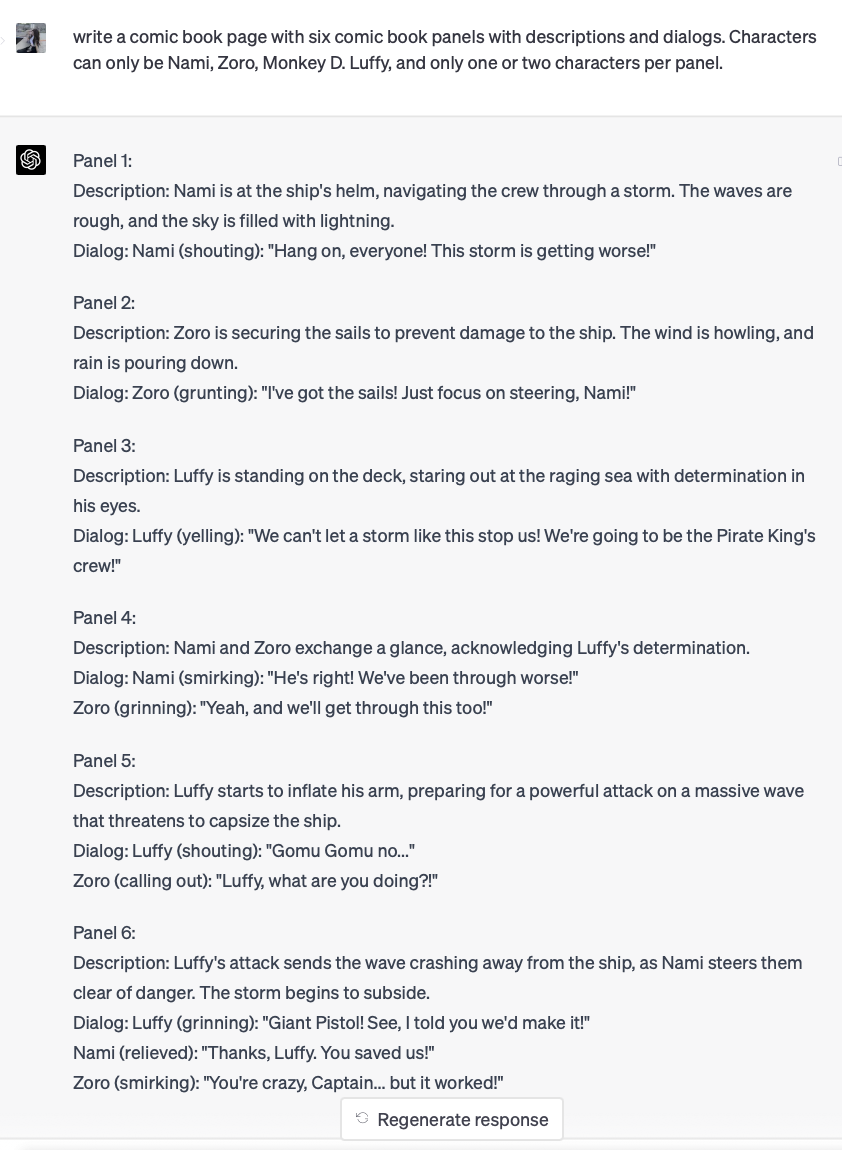}

\subsection{Sample Comic Generated using fine-tuned Stable Diffusion}
\includegraphics[width=\textwidth]{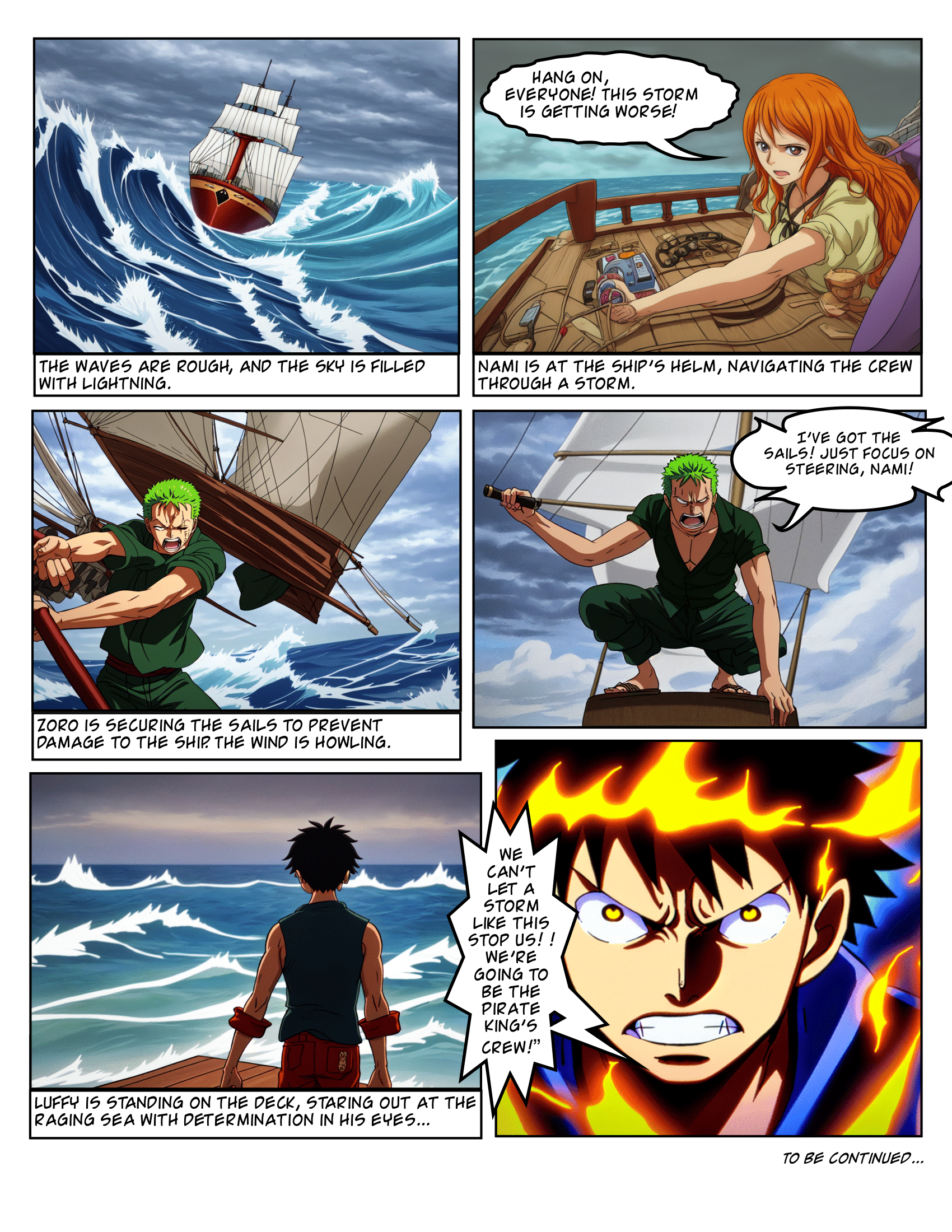}

\bibliography{sample}

\begin{thebibliography}{24}
\providecommand{\natexlab}[1]{#1}
\providecommand{\url}[1]{\texttt{#1}}
\expandafter\ifx\csname urlstyle\endcsname\relax
  \providecommand{\doi}[1]{doi: #1}\else
  \providecommand{\doi}{doi: \begingroup \urlstyle{rm}\Url}\fi

\bibitem[pyp()]{pypi}
Sentence-transformers.
\newblock URL \url{https://pypi.org/project/sentence-transformers/}.

\bibitem[Chen()]{chen}
M.~Chen.
\newblock How to validate openai gpt model performance with text summarization
  ...
\newblock URL
  \url{https://towardsdatascience.com/how-to-validate-openai-gpt-model-performance-with-text-summarization-298978fea764}.

\bibitem[CompVis()]{compvis}
CompVis.
\newblock Compvis/stable-diffusion: A latent text-to-image diffusion model.
\newblock URL \url{https://github.com/CompVis/stable-diffusion}.

\bibitem[Devlin et~al.(2019)Devlin, Chang, Lee, and Toutanova]{devlin2019bert}
J.~Devlin, M.-W. Chang, K.~Lee, and K.~Toutanova.
\newblock Bert: Pre-training of deep bidirectional transformers for language
  understanding, 2019.

\bibitem[DrBob2142()]{drbob2142}
DrBob2142.
\newblock Drbob2142/midnight\_mixes · hugging face.
\newblock URL \url{https://huggingface.co/DrBob2142/Midnight\_Mixes}.

\bibitem[El-Nouby et~al.(2019)El-Nouby, Zhai, Taylor, and
  Susskind]{elnouby2019skipclip}
A.~El-Nouby, S.~Zhai, G.~W. Taylor, and J.~M. Susskind.
\newblock Skip-clip: Self-supervised spatiotemporal representation learning by
  future clip order ranking, 2019.

\bibitem[fandom()]{fandom}
fandom.
\newblock One piece wiki.
\newblock URL \url{https://onepiece.fandom.com/wiki/One\_Piece\_Wiki}.

\bibitem[Gal et~al.(2022)Gal, Alaluf, Atzmon, Patashnik, Bermano, Chechik, and
  Cohen-Or]{gal2022image}
R.~Gal, Y.~Alaluf, Y.~Atzmon, O.~Patashnik, A.~H. Bermano, G.~Chechik, and
  D.~Cohen-Or.
\newblock An image is worth one word: Personalizing text-to-image generation
  using textual inversion, 2022.

\bibitem[hakurei()]{hakurei}
hakurei.
\newblock Hakurei/waifu-diffusion-v1-4 · hugging face.
\newblock URL \url{https://huggingface.co/hakurei/waifu-diffusion-v1-4}.

\bibitem[Heusel et~al.(2018)Heusel, Ramsauer, Unterthiner, Nessler, and
  Hochreiter]{heusel2018gans}
M.~Heusel, H.~Ramsauer, T.~Unterthiner, B.~Nessler, and S.~Hochreiter.
\newblock Gans trained by a two time-scale update rule converge to a local nash
  equilibrium, 2018.

\bibitem[Hu et~al.(2021)Hu, Shen, Wallis, Allen-Zhu, Li, Wang, Wang, and
  Chen]{hu2021lora}
E.~J. Hu, Y.~Shen, P.~Wallis, Z.~Allen-Zhu, Y.~Li, S.~Wang, L.~Wang, and
  W.~Chen.
\newblock Lora: Low-rank adaptation of large language models, 2021.

\bibitem[Huawei(2019)]{huawei}
Huawei.
\newblock Huawei presents ‘unfinished symphony’.
\newblock 2019.
\newblock URL
  \url{https://consumer.huawei.com/uk/campaign/unfinishedsymphony/}.

\bibitem[Katz(2022)]{Katz2022}
L.~Katz.
\newblock Ai drew this gorgeous comics series. you’d never know it, Dec 2022.
\newblock URL
  \url{https://www.cnet.com/culture/ai-drew-this-gorgeous-comics-series-youd-never-know-it/}.

\bibitem[Kingma and Welling(2022)]{kingma2022autoencoding}
D.~P. Kingma and M.~Welling.
\newblock Auto-encoding variational bayes, 2022.

\bibitem[lllyasviel()]{lllyasviel}
lllyasviel.
\newblock Lllyasviel/sd-controlnet-canny · hugging face.
\newblock URL \url{https://huggingface.co/lllyasviel/sd-controlnet-canny}.

\bibitem[Lykon()]{lykon}
Lykon.
\newblock One piece (wano saga) style lora | stable diffusion lora | civitai.
\newblock URL
  \url{https://civitai.com/models/4219/one-piece-wano-saga-style-lora}.

\bibitem[midjourney()]{midjourney}
midjourney.
\newblock URL \url{https://www.midjourney.com/}.

\bibitem[OpenAI(2023)]{openai2023gpt4}
OpenAI.
\newblock Gpt-4 technical report, 2023.

\bibitem[Ruiz et~al.(2023)Ruiz, Li, Jampani, Pritch, Rubinstein, and
  Aberman]{ruiz2023dreambooth}
N.~Ruiz, Y.~Li, V.~Jampani, Y.~Pritch, M.~Rubinstein, and K.~Aberman.
\newblock Dreambooth: Fine tuning text-to-image diffusion models for
  subject-driven generation, 2023.

\bibitem[Shen et~al.(2023)Shen, Song, Tan, Li, Lu, and
  Zhuang]{shen2023hugginggpt}
Y.~Shen, K.~Song, X.~Tan, D.~Li, W.~Lu, and Y.~Zhuang.
\newblock Hugginggpt: Solving ai tasks with chatgpt and its friends in
  huggingface, 2023.

\bibitem[Song and Jin()]{song_jin}
Z.~Song and Z.~Jin.
\newblock Zorinaaaaa/csc2516-project.
\newblock URL \url{https://github.com/Zorinaaaaa/csc2516-project}.

\bibitem[Wang et~al.(2004)Wang, Bovik, Sheikh, and Simoncelli]{1284395}
Z.~Wang, A.~Bovik, H.~Sheikh, and E.~Simoncelli.
\newblock Image quality assessment: from error visibility to structural
  similarity.
\newblock \emph{IEEE Transactions on Image Processing}, 13\penalty0
  (4):\penalty0 600--612, 2004.
\newblock \doi{10.1109/TIP.2003.819861}.

\bibitem[Wu et~al.(2023)Wu, Yin, Qi, Wang, Tang, and Duan]{wu2023visual}
C.~Wu, S.~Yin, W.~Qi, X.~Wang, Z.~Tang, and N.~Duan.
\newblock Visual chatgpt: Talking, drawing and editing with visual foundation
  models, 2023.

\bibitem[Zhang and Agrawala(2023)]{zhang2023adding}
L.~Zhang and M.~Agrawala.
\newblock Adding conditional control to text-to-image diffusion models, 2023.

\end{thebibliography}
\bibliographystyle{abbrvnat}

\end{document}